\pdfoutput=1

\documentclass[11pt]{article}

\usepackage{emnlp2021}

\usepackage{times}
\usepackage{latexsym}
\usepackage{multirow}
\usepackage{colortbl}
\usepackage{booktabs}
\usepackage{amssymb}
\usepackage{amsmath}
\usepackage{amsthm}
\usepackage{graphicx}
\usepackage[T1]{fontenc}

\usepackage[utf8]{inputenc}

\usepackage{microtype}

\definecolor{mygray}{gray}{.9}
\newcommand{\tabincell}[2]{\begin{tabular}{@{}#1@{}}#2\end{tabular}}
\title{Low-Resource Dialogue Summarization with Domain-Agnostic Multi-Source Pretraining}

\author{Yicheng Zou$^{1,2}$,\ \ Bolin Zhu$^2$,\ \ Xingwu Hu$^2$,\ \  Tao Gui$^1$\thanks{{ }{ }Corresponding authors.}\ \ ,\ \  Qi Zhang$^2$$^*$\\
  $^1$Institute of Modern Languages and Linguistics, Fudan University \\
  $^2$Shanghai Key Laboratory of Intelligent Information Processing, Fudan University\\
  $^2$School of Computer Science, Fudan University\\
Shanghai, China\\
  \texttt{\{yczou18,blzhu20,xwhu20,tgui,qz\}@fudan.edu.cn}}

\begin{document}
\maketitle
\begin{abstract}
With the rapid increase in the volume of dialogue data from daily life, there is a growing demand for dialogue summarization. Unfortunately, training a large summarization model is generally infeasible due to the inadequacy of dialogue data with annotated summaries. Most existing works for low-resource dialogue summarization directly pretrain models in other domains, e.g., the news domain, but they generally neglect the huge difference between dialogues and conventional articles. To bridge the gap between out-of-domain pretraining and in-domain fine-tuning, in this work, we propose a multi-source pretraining paradigm to better leverage the external summary data. Specifically, we exploit large-scale in-domain non-summary data to separately pretrain the dialogue encoder and the summary decoder. The combined encoder-decoder model is then pretrained on the out-of-domain summary data using adversarial critics, aiming to facilitate domain-agnostic summarization. The experimental results on two public datasets show that with only limited training data, our approach achieves competitive performance and generalizes well in different dialogue scenarios.
\end{abstract}

\section{Introduction}

With the explosion in the quantity of dialogue data from the Internet and daily life, there is growing interest in automatic dialogue summarization for various scenarios and applications, such as email threads, meetings, customer service, and online chats \cite{murray2008summarizing,shang2018unsupervised,liu2019automatic,zou2020unsupervised,zou2020topic}. Unfortunately, creating large-scale dialogue datasets with annotated summaries is costly and labor-intensive, which makes it difficult to build and train large summarization models using adequate supervision signals, especially in a new domain. Hence, it is necessary to develop models for dialogue summarization in low-resource settings, where only limited or even no training examples are available.

Recently, domain adaptation approaches with large-scale pretraining have attracted much attention in low-resource summarization \cite{wang2019exploring,yang2020ted,zhang2020pegasus}. A similar strategy is used in dialogues, whereby external summary data from other domains, e.g., the CNN/Dailymail news dataset \cite{hermann2015teaching}, are introduced for model pretraining prior to the final fine-tuning on low-resource dialogue summaries. Recent works  \cite{gliwa2019samsum,zhu2020hierarchical,joshi2020dr} have also reported the effectiveness of pretrained summarizers for different kinds of dialogue scenarios, such as chat logs and medical conversations. 

However, dialogue summary data has several inherent and significant differences from conventional articles in terms of text styles and summary structures. (i) Dialogues generally contain multiple participants who have distinct characteristics. (ii) Rather than the formal expressions found in news documents, dialogues often comprise utterances with informal or ungrammatical phrases. (iii) The structure of a dialogue summary, including length and the level of abstraction, is quite different from that in other domains \cite{zhu2020hierarchical}, e.g., CNN/Dailymail. Thus, considering the huge difference between dialogues and general documents, direct finetuning on dialogue summaries is not ideal when using a model pretrained from other domains.

To better leverage summary data from domains such as news or scientific articles, in this work, we introduce a novel pretraining paradigm called domain-agnostic multi-source pretraining (DAMS) to summarize dialogues in a low-resource setting. We postulate that the pretraining of dialogue summarization could be decomposed into three procedures: the pretraining of encoder, decoder, and the combined encoder-decoder model. Specifically, the dialogue encoder is pretrained on large-scale unannotated dialogues to learn the way of dialogue modeling and understanding. The summary decoder is pretrained on large-scale summary-like short texts to learn a language model in the style of the dialogue summaries. Furthermore, the encoder and decoder are combined and pretrained on external summary data to go through an integral process of summarization. The above pretraining processes from the three sources are performed simultaneously. By this means, DAMS exploits large-scale non-summary data in the same domain to narrow the gap between pretraining and fine-tuning. Additionally, adversarial critics are used to capture the features shared between dialogues and general documents, and to learn to perform domain-agnostic summarization.

We conducted experiments on two public dialogue summary datasets, namely SAMSum \cite{gliwa2019samsum} and ADSC \cite{misra2015using}. Pretraining was conducted on datasets from multiple sources, including dialogue corpora, daily-life short text corpora, and text summarization datasets from the news domain. The experimental results show that with only limited training data of dialogue summaries, our approach achieved competitive performance and showed a promising ability for generalizing different dialogue scenarios. Our codes and datasets are publicly available\footnote{\url{https://github.com/RowitZou/DAMS}}. 

In summary, our contributions are three-fold: 1) We explore the task of dialogue summarization in a low-resource setting with the usage of external multi-source corpora. 2) A novel pretraining strategy is designed to bridge the gap between out-of-domain pretraining and in-domain fine-tuning for domain-agnostic summarization. 3) Comprehensive studies on two datasets show the effectiveness of our method in various aspects.

\section{Related Work}
\subsection{Dialogue Summarization}
Dialogue summarization is a challenging and valuable task that receives much attention in recent years. Different from studies on conventional documents like news or reviews \cite{see2017get,narayan2018don,chu2019meansum}, dialogue summarization is investigated in multi-party interactions such as mail threads \cite{rambow2004summarizing}, meetings \cite{gillick2009global,shang2018unsupervised,zhong2021qmsum}, telephone conversation records \cite{zechner2001automatic,gurevych2004semantic}, and daily chats \cite{gliwa2019samsum,zhao2020improving}. Most of these approaches share a similar prerequisite: a decent labeled training dataset with annotated summaries. Nevertheless, creating a large-scale dialogue summary dataset is very expensive and labor-intensive, which makes the traditional methods hard to apply in real-world applications, especially when only limited or even no training signals are available. In this work, we explore dialogue summarization in a low-resource setting, and leverage external large-scale corpora to facilitate the task, which is applicable to most dialogue scenarios.

\begin{figure*}
\centering
  \includegraphics[width=6.3in]{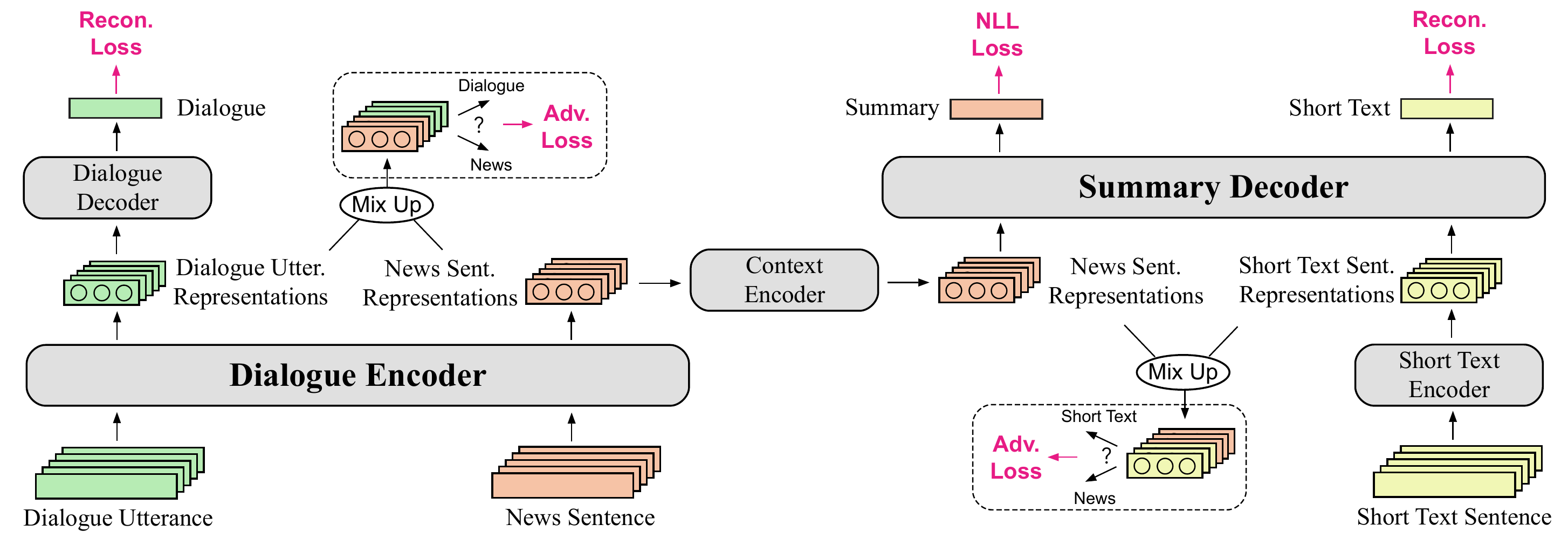}
  \caption{The overall architecture of DAMS. The multi-source pretraining includes: (i) encoder pretraining using dialogues (green); (ii) decoder pretraining using short texts (yellow); (iii) Joint pretraining using general articles with corresponding summaries (orange).} \label{fig:model}
\end{figure*}

\subsection{Domain Adaptation for Summarization}
Since texts and their summaries across diverse domains might share similarities and benefit from each other, domain adaptation for text summarization has attracted much research interest recently \cite{hua2017pilot,wang2019exploring,zhang2020pegasus,yang2020ted,yu2021adaptsum}. Most existing works perform pretraining on large-scale out-of-domain datasets and then adapt to the in-domain summary data. For dialogue summarization, although it is more ideal to perform adaptation from a source dialogue domain to a target dialogue domain \cite{sandu2010domain,wang2013domain}, unfortunately, the inadequacy of dialogue summary data makes it infeasible to directly train a large summarization model on the source data in an end-to-end manner. Recently, a couple of works have leveraged large-scale summary data that is more distinct from the dialogue domain, e.g., the news domain, to facilitate dialogue summarization \cite{gliwa2019samsum,zhu2020hierarchical,joshi2020dr}. However, the huge gap between dialogues and general articles is barely noticed. Yu et al. \shortcite{yu2021adaptsum} conducted pretraining on the news summary data and the dialogue non-summary data simultaneously, but the two different tasks share a single decoder, which might confuse the model about the knowledge that it learns. To better leverage the out-of-domain summary data and the in-domain non-summary data, we explore the domain-agnostic summarization. It is supported by a multi-source pretraining paradigm with adversarial learning, where the encoder and the decoder are separately pretrained on the in-domain non-summary data and combinedly pretrained on the out-of-domain summary data, aiming to narrow the gap between pretraining and fine-tuning.

\section{Methodology}

In this section, we detail the low-resource dialogue summarization under the domain-agnostic multi-source pretraining (DAMS). It consists of three pretraining objectives: two reconstruction losses with denoising auto-encoders that learn dialogue modeling and summary-like text generation; a sequence-to-sequence (seq2seq) training objective with the combined dialogue encoder and summary decoder that learns abstractive summarization. Additionally, two adversarial critics are attached to the encoder's output representations and the decoder's input representations, learning to perform domain-agnostic summarization. The overall framework is illustrated in Figure \ref{fig:model}.

\subsection{Multi-Source Pretraining}

Despite the considerable amount of summary data in other domains such as news and scientific articles, adaptation to the dialogue domain is not easy due to the huge difference between dialogues and conventional articles. To address this issue, we postulate that abstractive dialogue summarization could be decomposed into three procedures: (i) Dialogue modeling for understanding dialogue semantics and capturing dialogue characteristics; (ii) Saliency estimation based on learned representations to identify the important parts of input contents; (iii) Generating a summary grounded on the salient information with a certain style or structure. Although the limited dialogue summary data is inadequate to train the three procedures jointly, each one of them, fortunately, could be well handled by separate pretraining with large-scale corpora from different sources. Specifically, dialogue modeling can benefit from the usage of large-scale unannotated dialogues. The external news summary data may contribute to the process of saliency estimation. A language model trained on daily-life short texts can generate discourses with the style of dialogue summaries, rather than formal expressions in news or scientific articles.

{\bf Pretraining of dialogue modeling.} 
Inspired by recent large-scale pretraining models \cite{devlin2019bert,zhang2019hibert,lewis-etal-2020-bart}, we exploit the framework of denoising auto-encoding (DAE) \cite{vincent2008extracting} to extract robust features to compose dialogue representations. Formally, we denote each dialogue as an utterance sequence $D=\{u_1,u_2...,u_n\}$. To incorporate multi-party information, we add the name of the speaker at the beginning of each utterance. Then, we tokenize utterances into word sequences, denoted as $u_i=\{w_{i1},..,w_{im}\}$, where $w_{ij}$ is the $j$-th word in the sequence of $u_i$. For noise addition, we randomly mask 15\% of the tokens in each utterance with a special $\mathrm{[MASK]}$ token similar to BERT\footnote{In practise, we keep 20\% of the utterances unchanged. The purpose of this is to bias the representation towards the actual observed utterance.} \cite{devlin2019bert}. The purpose of noise addition is to encourage DAE to reconstruct the original utterances for robust representation learning. 

In this work, we employ Transformer with multi-head attentions \cite{vaswani2017attention} as the basic encoder and decoder of DAE. Before inputting word sequences into the encoder, we concatenate a special token $\mathrm{[CLS]}$ in front of each sequence similar to BERT. The final hidden state corresponding to this token is used as the aggregate sequence representation for utterance reconstruction. Formally, we transform the modified noisy sequence $\widetilde{u}_i=\{w^{cls}_{i},w_{i1}...,w_{im}\}$ into a sequence of hidden vectors by a Transformer encoder:
\begin{equation}
    [\mathbf{h}^{cls}_{i},\mathbf{h}_{i1},...,\mathbf{h}_{im}] = \mathrm{TF}_{\theta^d_e}([\mathbf{e}^{cls}_{i},\mathbf{e}_{i1},...,\mathbf{e}_{im}]),
    \label{eq:1}
\end{equation}
where $\mathbf{e}_{ij}$ is the embedding of the $j$-th word $w_{ij}$ in the word sequence, while $w^{cls}_{i},\mathbf{e}^{cls}_{i}$ represent $\mathrm{[CLS]}$ and its embedding.

The decoder is an auto-regressive model that recovers the original utterance conditioned on the input representation $\mathbf{h}^{cls}_{i}$. Here, we use a Transformer decoder with masked attention that conditions by adding $\mathbf{h}^{cls}_{i}$ to each input embedding. This is a Transformer variant that removes the decoder-encoder attention layer. Formally, the generation probability is defined by:
\begin{equation}
    P(\hat{w}_{ij}|\hat{w}_{i(1:j-1)};\widetilde{u}_i)= \mathrm{TF}_{\theta^d_g}([\hat{\mathbf{e}}_{i(1:j-1)}];\mathbf{h}^{cls}_{i}),
    \label{eq.2}
\end{equation}
where $\hat{\mathbf{e}}_{ij}$ denotes the embedding of the predicted word $\hat{w}_{ij}$ at the decoding step $j$. Notably, the decoder applies utterance representations $\mathbf{h}^{cls}_{i}$ as memories instead of using word-level attention or copy mechanism. It encourages all semantics to be captured in $\mathbf{h}^{cls}_{i}$. In Section \ref{method_dis}, we give a further discussion about why we do not choose the word-level cross-attention. Finally, we use the original utterance $u_i$ as a gold reference to train the DAE for utterance reconstruction on large-scale dialogue corpora, paving the way to dialogue modeling for the downstream summarization task:
\begin{equation}
    \mathcal{L}_{rec} = -\sum \nolimits_i \sum \nolimits_{j=1}^m \mathrm{log}P(\hat{w}_{ij}|\hat{w}_{i(1:j-1)};\widetilde{u}_i).
\end{equation}

{\bf Pretraining of summary language modeling.} We use the similar strategy as in dialogue pretraining to learn a summary language model. Here, we introduce the external corpora that contain daily-life short texts or stories, e.g., BooksCorpus \cite{zhu2015aligning}, to train the decoder to generate texts in the style of dialogue summaries. We truncate long documents into text pieces to form training samples, each one of which includes several consecutive sentences. We also add noise to these text pieces and train a DAE to recover them. Specifically, given the sentence sequence of a training sample $S=\{s_1,s_2,...,s_n\}$, we use the same noise addition strategy as for dialogues to construct noisy sentences, and encode them into hidden vectors by a Transformer encoder $\mathrm{TF}_{\theta^s_e}$ similar to Eq.\ref{eq:1}. 

The generation process, however, is different from that of utterance reconstruction. Since a summary might contain more than one sentence, we should encourage the decoder to generate all sentences of $S$ sequentially to simulate the process of summary generation. Hence, to further capture the global semantic dependency between sentences, we use another Transformer encoder to hierarchically fuse context information:
\begin{equation}
    [\hat{\mathbf{h}}^{cls}_{1},\hat{\mathbf{h}}^{cls}_{2},...,\hat{\mathbf{h}}^{cls}_{n}] = \mathrm{TF}_{\theta^s_h}([\mathbf{h}^{cls}_{1},\mathbf{h}^{cls}_{2},...,\mathbf{h}^{cls}_{n}]).
    \label{eq.4}
\end{equation}
Here, all sentence representations derived from $\mathrm{[CLS]}$ tokens are fed into the hierarchical encoder for information interaction. The output vectors are then used as memories for decoder-encoder attention in a classic Transformer decoder to recover $S$. The generation probability is:
\begin{align}
    \label{eq.5}
    &\hat{\mathbf{H}}^{cls} = [\hat{\mathbf{h}}^{cls}_{1},\hat{\mathbf{h}}^{cls}_{2},...,\hat{\mathbf{h}}^{cls}_{n}], \\ \nonumber
    &P(\hat{w}_k|\hat{w}_{1:k-1};\widetilde{S})= \mathrm{TF}_{\theta^s_g}([\hat{\mathbf{e}}_{1:k-1}];\hat{\mathbf{H}}^{cls}),
\end{align}
where $\widetilde{S}$ represents the noisy text piece and $\hat{w}_k,\hat{\mathbf{e}}_k$ denote the $k$-th predicted word and its embedding. The difference between Eq.\ref{eq.2} and Eq.\ref{eq.5} is that the former reconstructs a single utterance, while the latter predicts the entire text sample. Finally, we train the language model conditioned on $\widetilde{S}$ as:
\begin{equation}
    \mathcal{L}_{gen} = -\sum \nolimits_k \mathrm{log}P(\hat{w}_{k}|\hat{w}_{1:k-1};\widetilde{S}).
\end{equation}

{\bf Pretraining of abstractive summarization.} In order to pretrain end-to-end summary generation, we bridge the dialogue encoder $\mathrm{TF}_{\theta^d_e}$ with the summary decoder $\mathrm{TF}_{\theta^s_g}$ using a context encoder $\mathrm{TF}_{\theta^b_h}$. $\mathrm{TF}_{\theta^b_h}$ has the same architecture as in Eq.\ref{eq.4}. Then, we input sentences of a document into $\mathrm{TF}_{\theta^d_e}$ and get a predicted summary from $\mathrm{TF}_{\theta^s_g}$, training the model with the following objective:
\begin{equation}
    \mathcal{L}_{summ} = -\sum \nolimits_k \mathrm{log}P(\hat{w}_{k}|\hat{w}_{1:k-1};D_{s}),
    \label{eq:7}
\end{equation}
where $D_s$ is the document. $\hat{w}_k$ represents the $k$-th word in the predicted summary. Here, we reuse $\mathrm{TF}_{\theta^d_e}$ and $\mathrm{TF}_{\theta^s_g}$ for abstractive summarization, and its purpose is to bridge the gap between separate pretraining on multi-source texts and joint fine-tuning on dialogue summaries. By an integral process of text summarization, the combined encoder-decoder model learns to capture salient information from sentence (or utterance) representations and generate summaries accordingly.

\subsection{Domain-Agnostic Summarization with Adversarial Learning}
Ideally, the DAE learns a high-level latent {\em content} conveyed in representations, disentangled from their original attributes, e.g., styles of informal dialogue utterances and formal news sentences, adapting the way of saliency estimation and summary generation to the dialogue domain. However, models often learn domain-specific features, making it difficult to generalize in a new domain \cite{peng2019domain}. To address this issue, inspired by recent works of adversarial summary generation \cite{liu2018generative,rekabdar2019generative}, we add an adversarial discriminator (critic) that learns to identify the domain of each representation, and use a gradient reversal mechanism \cite{ganin2015unsupervised} to ensure that the feature distributions over different domains are made similar (as indistinguishable as possible for the discriminator), thus resulting in the domain-invariant features and encouraging the summarizer to only focus on content rather than domain-specific attributes.

Here, we add two adversarial critics $\mathrm{D}_e, \mathrm{D}_g$ on the output vectors of $\mathrm{TF}_{\theta^d_e}$ and the input vectors of $\mathrm{TF}_{\theta^s_g}$, respectively (see Figure \ref{fig:model}). The former classifies output vectors as dialogue utterances or news sentences, and the latter tries to distinguish news articles from short texts. The adversarial critic is a simple binary classifier with a multi-layer perceptron and a sigmoid activator trained by a logistic loss function, denoted as $\mathcal{L}^D_{e},\mathcal{L}^D_g$ for $\mathrm{D}_e$ and $\mathrm{D}_g$, respectively. Finally, we combine all pretraining losses and adversarial signals to jointly train the model, where $\alpha$ is a hyper-parameter to adjust the loss proportion:
\begin{equation}
\mathcal{L}=\mathcal{L}_{rec} + \mathcal{L}_{gen} + \mathcal{L}_{summ} + \alpha(\mathcal{L}^D_{e} + \mathcal{L}^D_{g}). 
\end{equation}

\subsection{Fine-tuning on Dialogue Summaries}
After multi-source pretraining, we further stack $\mathrm{TF}_{\theta^d_e}$, $\mathrm{TF}_{\theta^b_h}$, and $\mathrm{TF}_{\theta^s_g}$ for joint fine-tuning on the dialogue summary dataset. The learning objective is similar to Eq.\ref{eq:7}. Notably, the three modules are fully trained by appropriate data from multiple sources, leading to a higher convergence speed on the target dialogue summaries (see details in Section \ref{sec:discussion}), which requires fewer training data points to achieve a competitive performance. 

\subsection{Discussion of the Encoder-Decoder Connection Strategy}
\label{method_dis}
The encoder-decoder cross attention for encoding the context information is widely used in transformer-based architectures. Large-scale pretraining models for the summarization task, e.g., BART \cite{lewis-etal-2020-bart}, generally exploit token-level attention to integrate the document context. In this work, we have tried keeping the traditional token-level cross attention in the proposed architecture to directly connect the dialogue encoder and the summary decoder. However, we find that it is difficult to disentangle the encoder and the decoder for separate pretraining. It is also hard to add adversarial critics to token-level representations involved in the cross attention to learn domain-invariant features. Considering the above limitations, we use an embedding concatenation strategy in the dialogue decoder $\mathrm{TF}_{\theta^d_g}$ as a DAE to learn utterance representations. The summary decoder $\mathrm{TF}_{\theta^s_g}$ still has the cross attention, but keys and values are sentence representations from the context encoder $\mathrm{TF}_{\theta^b_h}$ instead of token representations from the dialogue encoder $\mathrm{TF}_{\theta^d_e}$. Here, $\mathrm{TF}_{\theta^b_h}$ bridges the dialogue encoder and the summary decoder. It not only captures the context information of sentences (utterances), but also derives sentence-level representations that are applicable for domain identification in adversarial learning. Nevertheless, the abandonment of token-level attention will inevitably affect the fine-grained information integration. In terms of how to keep the token-level cross attention in DAMS, we leave it as a future work for open discussions. 

\begin{table}[t!]
\footnotesize
\begin{center}
\setlength{\tabcolsep}{2mm}{
\begin{tabular}{cccccc}
\toprule[1pt]
 \multirow{2}{*}{{\bf Dataset}} &\multirow{2}{*}{{\bf Split}} & {\bf \# of} & {\bf Avg.}& {\bf Avg.}& {\bf Ref.}\\
& & {\bf dial.}& {\bf words}& {\bf turns}&{\bf length}\\
\midrule
\multirow{3}{*}{SAMSum} & Train & 14,732&120.26 & 11.13 & 22.81 \\
& Dev.& 818&117.46 & 10.72 & 22.80\\
& Test& 819&122.71 & 11.24 & 22.47\\
\midrule
ADSC & All & 45 & 370.44 & 7.51 & 101.99 \\
\bottomrule[1pt]
\end{tabular}}
\end{center}
\caption{\label{tb:data}Statistics of dialogue summary datasets. }
\end{table}

\section{Experimental Settings}
\subsection{Datasets}
Following the latest works \cite{zhao2020improving,feng2020incorporating}, we evaluate our method on two public dialogue summary datasets SAMSum \cite{gliwa2019samsum} and ADSC\footnote{Following \citet{feng2020incorporating}, we train the model using SAMSum corpus and perform zero-shot testing on ADSC.} \cite{misra2015using}. Statistics of the dialogue datasets is shown in Table \ref{tb:data}. SAMSum originally contains 14k
training examples. To simulate a low-resource scenario, we start from using the full training data, and gradually reduce the number of training examples by halving the training set. For multi-source pretraining, we use the following datasets. 

{\bf Dialogues.} We use Reddit Conversation Corpus \cite{dziri2019augmenting}\footnote{https://github.com/nouhadziri/THRED} for the pretaining of dialogue modeling. It contains about 15M context-response pairs for training, where each dialogue context consists of 3.5 utterances on average.

{\bf Short Texts.} We choose MSCOCO \cite{lin2014microsoft} and BookCorpus \cite{zhu2015aligning} to pretrain the summary language model. MSCOCO is a standard benchmark dataset for the image caption generation task, which contains over 120K images and 600K captions describing the prominent object/action in an image. Here, we only use captions to train the generator. BookCorpus is a large-scale corpus containing 11,038 free books from the Internet. We randomly truncate long documents into text pieces as training samples\footnote{Here, we use truncated sentence sequences in BookCorpus because we did not find other suitable corpora like MSCOCO. A real daily-life corpus with short-text summaries could be better for summary decoder pretraining.}. Each sample contains 1.5 sentences on average and we collect about 5M samples for training.

{\bf Summarization Corpus.} CNN/DailyMail \cite{hermann2015teaching}, Gigaword \cite{rush2015neural}, and NewsRoom \cite{grusky2018newsroom} are used as our external summary datasets for joint pretraining. All the three datasets are news articles or headlines with summaries from various news publications. We combine these datasets and the total training set consists of 5.6M samples.

\subsection{Comparison Methods}
For comparison, we select various baseline systems from previous literatures: the basic baseline {\bf Longest-3} \cite{gliwa2019samsum}, which selects the longest three utterances as a summary; Classic seq2seq models, including {\bf Seq2Seq+Attention} \cite{rush2015neural}, {\bf Transformer} \cite{vaswani2017attention}, and {\bf PGNet} \cite{see2017get}; A pipeline method {\bf FastRL} \cite{chen2018fast} and its variant {\bf FastRL Enhanced} \cite{gliwa2019samsum}, which first extracts salient sentences and then refines them; Convolution-based methods {\bf LightConv} \cite{wu2018pay} and {\bf DynamicConv} \cite{wu2018pay}; Methods based on graph neural networks, including {\bf D-HGN} \cite{feng2020incorporating} and {\bf TGDGA} \cite{zhao2020improving}; A seq2seq model {\bf BERT+TRF} \cite{liu2019text} that is equipped with pretrained LMs.

\subsection{Implementation Details}
At the pretraining stage, we mix up the datasets from multiple sources and keep dialogues, short texts, and news summaries in a percentage of 1:1:1. The total data points are around 15M. Since DAMS consists of Transformer encoders and decoders, it can be easily combined with pretrained LMs. Here, we use BERT \cite{devlin2019bert} as the utterance/sentence encoder $\mathrm{TF}_{\theta^d_e}$ and use a separate optimization strategy \cite{liu2019text} to alleviate the mismatch between BERT and other randomly initialized parameters. We apply Adam \cite{kingma2015adam} ($\beta_1$=0.9, $\beta_2$=0.999) with learning rate 1e-3 for BERT and 1e-2 for other parameters. All transformer blocks except BERT have 6 layers, 8 heads, 768 hidden units, and the hidden size for all feed-forward layers is 2048. Loss coefficient $\alpha$ is selected from $\{0.01,0.05,0.1,0.5\}$ to control adversarial signals, and we empirically find that $\alpha=0.1$ achieves the best performance on the validation set. The model is pretrained for 250,000 steps with 10,000 warm-up steps on 2 GeForce RTX 3090 GPUs. At the fine-tuning stage, we use the last pretraining checkpoint for fine-tuning on the SAMSum dataset. We continue to train the model for 50,000 steps with 1,000 warm-up steps using Adam ($\beta_1$ =0.9, $\beta_2$=0.999, learning rate=1e-3). During the inference time, summaries are decoded in a beam size of 3. The minimal summary length is set to 15 for SAMSum and 100 for ADSC, respectively. Checkpoints are saved and evaluated on the validation set every 2,000 steps. The best checkpoint trained on SAMSum is directly evaluated on ADSC to perform zero-shot testing.
\begin{table}[t!]
\footnotesize
\begin{center}
\setlength{\tabcolsep}{1.5mm}{
\begin{tabular}{|l|c|ccc|}
\hline
{\bf Model} & {\bf +News} & {\bf RG-1} & {\bf RG-2} & {\bf RG-L}\\
\hline
Longest-3 & -& 32.46 & 10.27 & 29.92 \\ 
Seq2Seq+Att & -& 29.35 & 15.90 & 28.16 \\ 
Transformer & -&37.27 & 18.44 & 32.73 \\ 
PGNet & -&40.08 & 15.28 & 36.63 \\
FastRL& -&40.96 & 17.18 & 39.05 \\ 
FastRL Enhanced & -& 41.95 &  18.06 & 39.23 \\ 
D-HGN & -&42.03 & 18.07 &39.56 \\ 
TGDGA & -&43.11 & 19.15 & 40.49 \\ 
BERT+TRF &- &39.90 & 17.01 & 39.12 \\
\hline
LightConv & \checkmark&40.29 & 17.28 & 36.81\\
DynamicConv& \checkmark&41.07& 17.11& 37.27\\ 
Transformer& \checkmark&42.37& 18.44 & 39.27 \\ 
PGNet & \checkmark&37.27 & 14.42& 34.36 \\ 
FastRL & \checkmark&41.03 & 16.93 & 39.05 \\
FastRL Enhanced& \checkmark& 41.87 & 17.47 & 39.53\\ 
BERT+TRF &\checkmark& 42.37& 17.59& 40.73\\
\hline
DAMS (w/o pretrain) &- &39.07 & 14.59 & 38.06 \\
DAMS & \checkmark &\bf 44.38 &\bf 19.98 &\bf 43.40 \\
\hline
\end{tabular}}
\end{center}
\caption{\label{tb:samsum}Results of ROUGE-1/2/L on the SAMSum corpus. {\bf +News} means whether the approach exploits external news summary data or not. }
\end{table}

\begin{table}[t!]
\footnotesize
\begin{center}
\setlength{\tabcolsep}{1.5mm}{
\begin{tabular}{|l|ccc|}
\hline
{\bf Model} & {\bf RG-1} & {\bf RG-2} & {\bf RG-L}\\
\hline
PGNet & 28.95& 5.34& 22.41\\ 
Transformer &27.13 &5.30 & 20.59\\ 
FastRL Enhanced & 30.00 & 4.87 & 22.27 \\
BERT+TRF* & 28.13& 4.63 &27.17\\
\hline
DAMS (w/o pretrain) & 28.17 &5.11& 27.09\\
DAMS* &\bf 31.29&\bf 5.53 & \bf 30.14\\
\hline
\end{tabular}}
\end{center}
\caption{\label{tb:adsc}Results of zero-shot testing on ADSC. Models marked with * use external news summary data.}
\end{table}
\section{Results and Analysis}
In this section, we show the main results of DAMS against other baselines for dialogue summarization, and probe the effectiveness of DAMS by explanatory experiments in various aspects. 

\subsection{Automatic Evaluation}
Table \ref{tb:samsum} and Table \ref{tb:adsc} show the results of automatic evaluation on the SAMSum and ADSC dataset. We evaluate summary quality using ROUGE F1 \cite{lin2004rouge}, including the unigram and bigram overlap (ROUGE-1, ROUGE-2) between system outputs and gold summaries, and the longest common subsequence (ROUGE-L). Some results are from the reported scores in previous literatures \cite{gliwa2019samsum,feng2020incorporating,zhao2020improving}.

In Table \ref{tb:samsum}, all baseline methods are categorized into two groups. The first group includes models that are directly trained on the SAMSum corpus, and methods in the second group benefit from external news summary data\footnote{For BERT+TRF, we pretrain it on our constructed news dataset. The other results are from Gliwa et al. \shortcite{gliwa2019samsum}, where models are trained on the train set of CNN/DailyMail joined together with the train set of SAMSum, and evaluated on the SAMSum test set. }. DAMS with full training data outperforms all baseline methods and is significantly different from BERT+TRF (+news) with $p<$ 0.05, which probes the superiority of the multi-source pretraining strategy for dialogue summarization against the general exploitation of news summary data. Without news data, DAMS might be inferior to seq2seq models like PGNet or BERT+TRF, because these models use word-level attentions or copy mechanisms, while DAMS focuses on sentence/utterance representations for domain-agnostic representation learning. We also observe that the inclusion of news summary data does not necessarily mean a better ROUGE score (PGNet, FastRL). One possible explanation is that these models learn domain-specific features and have difficulty adapting to the dialogue domain. By contrast, with news summary data, the performance of DAMS increases a lot, which validates that our method can successfully capture useful information from external corpora. Furthermore, we directly test models on the ADSC dataset to verify whether they can generalize well to a new scenario. From Table \ref{tb:adsc} we observe that DAMS performs best, indicating that our multi-source pretraining strategy enables well-pretrained parameters for the downstream dialogue summarization, which makes the model easier to adapt to other dialogue scenarios.

\begin{table}[t!]
\small
\begin{center}
\begin{tabular}{lcc}
\toprule[1pt]
\bf Methods &\bf Informativeness & \bf Fluency \\
\midrule[1pt]
PGNet & -0.128 & -0.246\\
Transformer & -0.210 & -0.119 \\
FastRL Enhanced & -0.103 & -0.052\\
BERT+TRF*& -0.037& \ 0.091 \\
DAMS*&\bf \ 0.088&\bf \ 0.102\\
\midrule
Gold & \ 0.390 & \ 0.224 \\
\bottomrule[1pt]
\end{tabular}
\end{center}
\caption{\label{human} Human evaluation with model ranking results. Models with * utilize external news summary data.}
\end{table}

\subsection{Human Evaluation}
Following \citet{narayan2018don}, we randomly sample 100 examples in the test set of SAMSum for human evaluation. Three volunteers are invited to compare summaries produced from 6 systems (including the gold summary). Given a dialogue and two summaries from two out of six systems, each volunteer should decide which summary is better on two dimensions: {\bf informativeness} (which summary captures more important information?) and {\bf fluency} (which summary is more fluent?). We collect judgments from three volunteers for each comparison to minimize the inter-human noise.

Table \ref{human} shows the system ranking results. Each score is calculated as the percentage of times the system is selected as best minus the percentage of times it is chosen as worst, ranging from -1 (worst) to 1 (best). {\bf Gold} unsurprisingly ranks best. For informativeness, volunteers exhibit more preference to DAMS. For fluency, models with pretraining (DAMS / BERT+TRF) produce more acceptable summaries. We carry out pairwise comparisons between systems (using a binomial two-tailed test; $p<$0.05). In terms of informativeness, DAMS is significantly different from all other systems. For fluency, pretrain-based systems significantly differ from other systems, and BERT+TRF is not significantly different from DAMS.

\begin{table}[t!]
\footnotesize
\begin{center}
\begin{tabular}{lccc}
\toprule[1pt]
\bf Methods &\bf RG-1 & \bf RG-2 & \bf RG-L\\
\midrule[1pt]
DAMS & 44.38 & 19.98 &  43.40\\
\ \ (w/o) $D_e$ & 42.29 & 18.33 & 41.28\\
\ \ (w/o) $D_g$ & 42.83& 18.48& 41.77\\
\ \ (w/o) $D_e$+$D_g$ &43.89 & 18.52& 42.09\\
\ \ (w/o) Dial. & 42.89& 18.17 & 41.60\\
\ \ (w/o) Short &43.01 & 18.65 & 41.71\\
\ \ (w/o) Summ. & 43.37& 17.98& 41.65\\
\bottomrule[1pt]
\end{tabular}
\end{center}
\caption{\label{tb:ablation} Ablation study of adversarial learning and multi-source pretraining. $D_e,D_g$ are two critics. {\bf Dial.}, {\bf Short}, and {\bf Summ.} denote corpora of dialogues, short texts, and news summaries, respectively.}
\end{table}

\begin{figure}
\centering
  \includegraphics[width=2.8in]{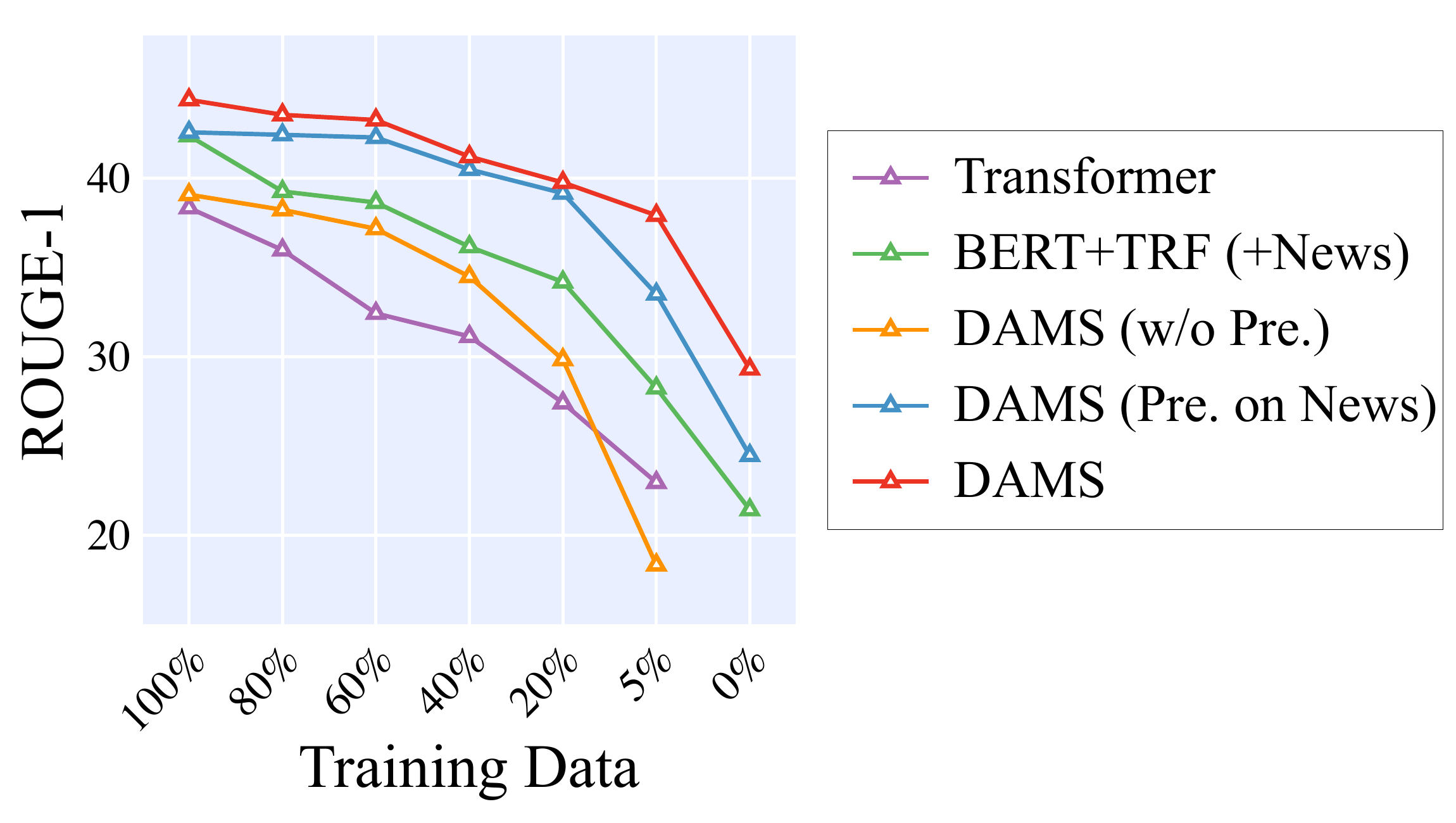}
  \caption{Model performance in low-resource settings.} \label{fig:low_res}
\end{figure}

\subsection{Analysis and Discussion}
\label{sec:discussion}We also perform qualitative analysis and discuss the effect of multi-source pretraining and adversarial learning with the following experiments. 

{\bf Ablation Study.} Table \ref{tb:ablation} shows the results of DAMS with different settings of adversarial critics and multi-source pretraining. We can see that the system suffers a performance degradation without the critic. It indicates that a domain-invariant representation is beneficial for downstream dialogue summarization. When any kind of external corpora is removed, the results drop a lot, which validates the effectiveness of multi-source pretraining.

\begin{figure}[t]
\centering
  \includegraphics[width=3.0in]{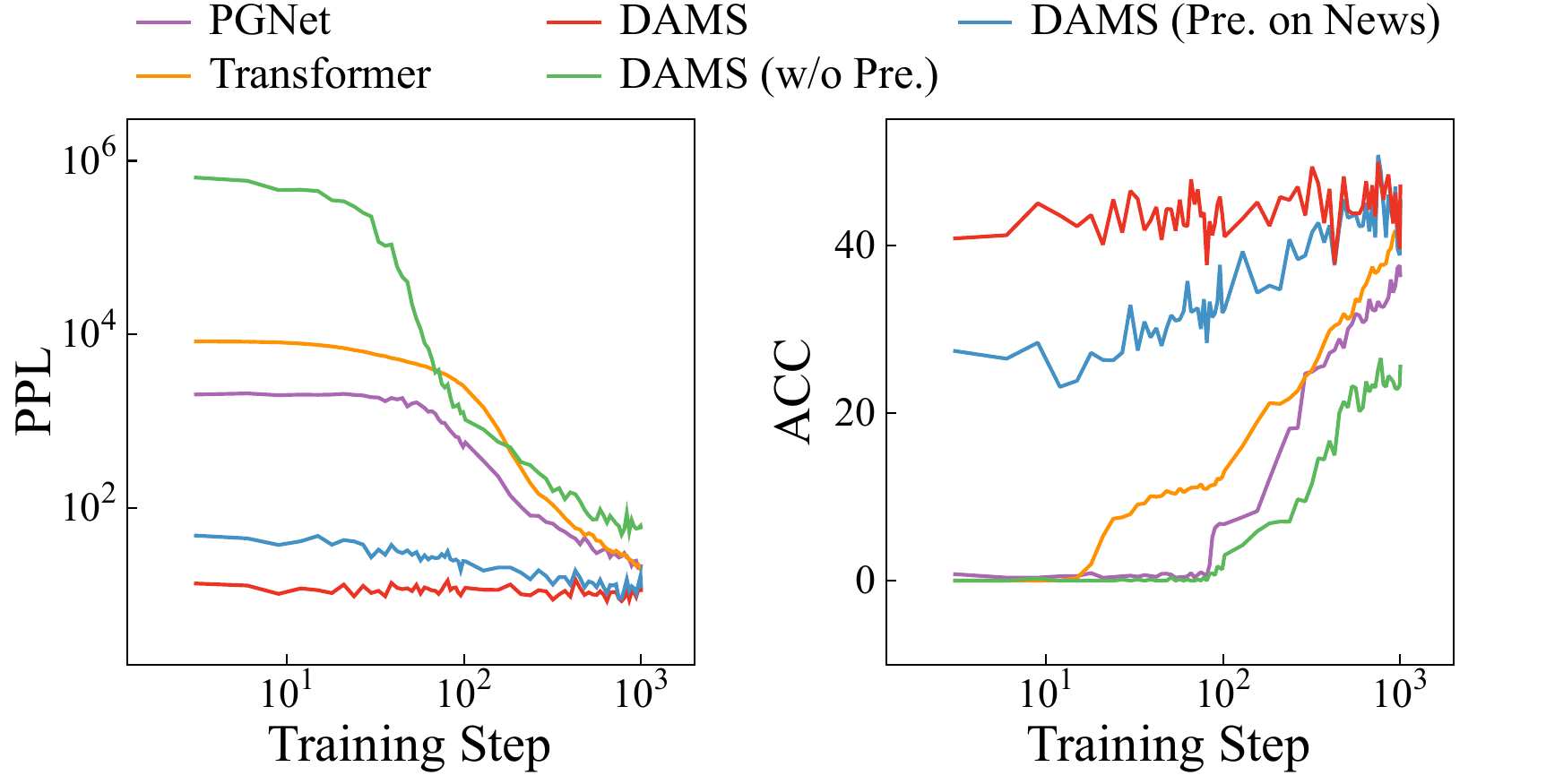}
  \caption{Fine-tuning logs of different models on the SAMSum dataset. PPL and ACC represent perplexity and word accuracy, respectively.} \label{fig:conv}
\end{figure}

{\bf Performance in Low-Resource Settings.} To analyze model performances in low-resource settings, we gradually reduce the number of training examples in the SAMSum corpus by halving the training set. We report the results of DAMS and two baseline methods (Transformer and BERT+TRF) with different percentages of training data in Figure \ref{fig:low_res}. We also report the performance of two variants of DAMS, without pretraining and only pretrained on the news summary data. Figure \ref{fig:low_res} shows a performance decline trend when the training data decreases continuously. We observe that with only limited SAMSum training data (40\% / 20\%), DAMS still achieves competitive results, while BERT+TRF (+News) suffers from a serious performance degradation. It indicates that DAMS has a promising ability of adapting news summaries to dialogue scenarios. Notably, using only 20\% of the training data, DAMS achieves a competitive performance against Transformer and DAMS (w/o Pre.) that use the full training data, which proves the effectiveness of exploiting external corpora. When the training set is cut to 5\% or even in a zero-shot setting, DAMS with multi-source pretraining shows a superior performance against all the other systems, including its variant DAMS (Pre. on News). It validates that our multi-source pretraining strategy is more applicable to dialogue summarization in a low-resource setting.

{\bf Convergence Rate.} In Figure \ref{fig:conv}, we demonstrate the fine-tuning logs of different models on the SAMSum dataset. The left figure shows the perplexity and the right figure shows the average word accuracy. Unsurprisingly, models that benefit from pretraining have better initialized parameters, leading to faster convergence. Equipped with the multi-source pretraining strategy, DAMS can perform better and even achieve a 40\% rate of word accuracy at the beginning of fine-tuning.

\begin{figure}[t]
\centering
  \includegraphics[width=3.0in]{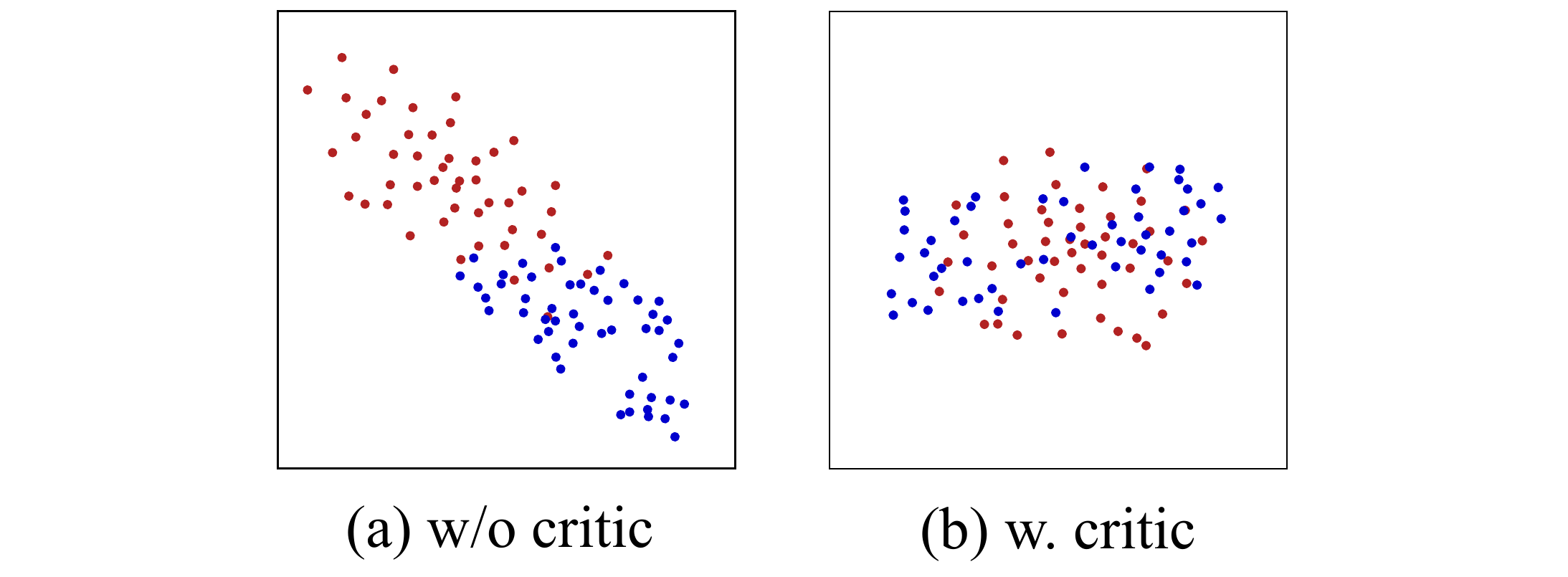}
  \caption{2-D visualizations of representations in the dialogue and news domain.} \label{fig:tsne}
\end{figure}

{\bf Domain-Agnostic Representations.} To verify the effectiveness of our adversarial strategy that can learn domain-agnostic features, we visualize the latent space of representations in 2-D using t-SNE \cite{van2008visualizing}, with and without the critic. In Figure \ref{fig:tsne}(a) where there is no critic, representations indeed show two separate clusters, while in Figure \ref{fig:tsne}(b), hidden vectors with adversarial signals are effectively merged into one region, resulting in domain-agnostic representations. It encourages the summarizer to focus on content rather than domain-specific attributes for better generalization from other domains to the dialogue domain.

{\bf Case Study.} Table \ref{tb:case} shows the system outputs of an exemplar dialogue. Texts with red color represent salient information in the dialogue, which is reflected in the gold summary. From the table we can see that DAMS can generate a summary that is more fluent and informative, which successfully captures critical information such as 'raining' and 'half an hour', composing a coherent discourse.

\section{Conclusion and Future Work}
In this paper, we propose a domain-agnostic multi-source pretraining paradigm for low-resource dialogue summarization, which exploits external large-scale corpora from multiple sources to facilitate dialogue modeling, summary language modeling, and abstractive summarization. The pretraining is conducted with adversarial signals to learn domain-agnostic summarization. The experimental results verify the effectiveness and generalization of our method in low-resource settings. Future directions are exploring how to keep the token-level cross attention in the multi-source pretraining strategy. In this way, we could adopt the strategy in the models with universal transformer architectures, e.g., BART, to benefit from large-scale pretraining language models.

\section*{Acknowledgements}
The authors wish to thank the anonymous reviewers for their helpful comments. This work was partially funded by China National Key R\&D Program (No. 2017YFB1002104), National Natural Science Foundation of China (No. 61976056, 62076069), Shanghai Municipal Science and Technology Major Project (No.2021SHZDZX0103).

\begin{table}[t]
\footnotesize
\begin{center}
\setlength{\tabcolsep}{1mm}{
\begin{tabular}{|c|l|}
\hline
\multirow{7}{*}{Dialogue} & Val\quad\ : it's raining!\\
& Candy: I know, just started...\\
& Val\quad\ : r we going? we will be wet\\
& Candy: maybe wait a little? see if stops\\
& Val\quad\ : ok. let's wait half h and than see\\
& Candy: god idea, I call u then\\
& Val\quad\ : great :)\\
\hline
\multirow{2}{*}{Gold} & It's {\color{purple}raining}, so {\color{purple}Val and Candy} will {\color{purple}wait} \\
& {\color{purple}half an hour} before they go.\\
\hline
\hline
\multirow{2}{*}{PGNet} & Val is learning to meet Val and Val will \\
& see a little.\\
\hline
\multirow{2}{*}{TRF}& Val and Val don't have any news. Val will \\
& call him because they got lost.\\
\hline
\multirow{2}{*}{\tabincell{c}{\tabincell{c}{DAMS \\ (w/o Pre.)}}}& {\color{purple}Candy and Val} are going to meet. Val will \\
& call Candy instead.\\
\hline
\multirow{2}{*}{\tabincell{c}{\tabincell{c}{BERT* \\ +TRF}}}& {\color{purple} Val and Candy} are going for a little, but \\
& they need to {\color{purple}wait half an hour}.\\
\hline
\multirow{2}{*}{DAMS*}& {\color{purple}Val and Candy} are going to {\color{purple}wait half} \\
& {\color{purple}an hour} to see if it's {\color{purple}raining}.\\
\hline
\end{tabular}}
\end{center}
\caption{\label{tb:case} System outputs of a dialogue example from the SAMSum test set. Systems marked with * utilize external news summary data.}
\end{table}

\bibliography{custom}
\bibliographystyle{acl_natbib}

\end{document}